\documentclass[letterpaper, 10 pt, conference]{ieeeconf}

\IEEEoverridecommandlockouts

\usepackage{wrapfig}

\usepackage{amsmath}
\usepackage{mathtools}
\usepackage{verbatim}
\usepackage{graphicx}
\graphicspath{ {Figures/} }
\usepackage{epstopdf}
\usepackage{tabularx}
\usepackage{booktabs} % \**rule table line package
\usepackage{multirow}
\usepackage{arydshln} % dashline package
\usepackage{url}
\usepackage{hyperref} % makes references hyperlinks
\usepackage{textcomp} % Symbol Expansion Package
\usepackage{float} % Adjust Composing package
\usepackage{color}
\usepackage{setspace}
\usepackage{fancyhdr}
\usepackage{longtable}
\usepackage[symbol]{footmisc}
\usepackage{pdfpages} 
\usepackage{rotating}
\usepackage[a-2b,mathxmp]{pdfx}

\title{\LARGE \bf
Efficient Concurrent Design of the Morphology of Unmanned Aerial Systems and their Collective-Search Behavior
}

\author{Chen Zeng$^{1}$, Prajit KrisshnaKumar$^{2}$, Jhoel Witter$^{3}$, and Souma Chowdhury$^{4}$% <-this % stops a space
\thanks{Authors $^{1}, ^{2}, ^{3}, ^{4}$ are with Mechanical and Aerospace Engineering,
        University at Buffalo, Buffalo, NY, USA
        {\tt\small \{czeng2, prajitkr, jhoelwit, soumacho\}@buffalo.edu}}%
\thanks{$^\dagger$ Corresponding Author, soumacho@buffalo.edu}
\thanks{This work was supported by the National Science Foundation (NSF) award CMMI 2048020. Any opinions, findings, conclusions, or recommendations expressed in this paper are those of the authors and do not necessarily reflect the views of the NSF.
}
\thanks {\copyright\space2022 IEEE. Personal use of this material is permitted. Permission from IEEE must be obtained for all other uses, in any current or future media, including reprinting/republishing this material for advertising or promotional purposes, creating new collective works, for resale or redistribution to servers or lists, or reuse of any copyrighted component of this work in other works.}
}

\begin{document}

\maketitle

% \thispagestyle{empty} % to disable page number display
% \pagestyle{empty}

% \thispagestyle{plain} % to display page number
% \pagestyle{plain}

%%%%%%%%%%%%%%%%%%%%%%%%%%%%%%%%%%%%%%%%%%%%%%%%%%%%%%%%%%%%%%%%%%%%%%
\begin{abstract}

The collective operation of robots, such as unmanned aerial vehicles (UAVs) operating as a team or swarm, is affected by their individual capabilities, which in turn is dependent on their physical design, aka morphology. However, with the exception of a few (albeit ad hoc) evolutionary robotics methods, there has been very little work on understanding the interplay of morphology and collective behavior. There is especially a lack of computational frameworks to concurrently search for the robot morphology and the hyper-parameters of their behavior model that jointly optimize the collective (team) performance. To address this gap, this paper proposes a new co-design framework. Here the exploding computational cost of an otherwise nested morphology/behavior co-design is effectively alleviated through the novel concept of ``talent" metrics; while also allowing significantly better solutions compared to the typically sub-optimal sequential morphology$\to$behavior design approach. This framework comprises four major steps: talent metrics selection, talent Pareto exploration (a multi-objective morphology optimization process), behavior optimization, and morphology finalization. This co-design concept is demonstrated by applying it to design UAVs that operate as a team to localize signal sources, e.g., in victim search and hazard localization. Here, the collective behavior is driven by a recently reported batch Bayesian search algorithm called Bayes-Swarm. Our case studies show that the outcome of co-design provides significantly higher success rates in signal source localization compared to a baseline design, across a variety of signal environments and teams with 6 to 15 UAVs. Moreover, this co-design process provides two orders of magnitude reduction in computing time compared to a projected nested design approach. % An application of the optimization of asynchronized Bayes-Swarm search algorithm, along with the corresponding UAV design, is presented. Three talent metrics are selected following distinct principles, the four-stage sequential design optimization procedures are completed with a frugal computing cost. The optimized final design outperforms a real-world-inspired baseline in robustness despite a slight disadvantage in search time.

\end{abstract}

%%%%%%%%%%%%%%%%%%%%%%%%%%%%%%%%%%%%%%%%%%%%%%%%%%%%%%%%%%%%%%%%%%%%%%

%%%%%%%%%%%%%%%%%%%%%%%%%%%%%%%%%%%%%%%%%%%%%%%%%%%%%%%%%%%%%%%%%%%%%%%%%%%%%%%%%%%%%%%%%%%%%%%%%%%%%%% %%%%%%%%%%%%%%%%%%%%%%%%%%%%%%%%%%%%%%%%%%%%%%%%%%%%%%%%%%%%%%%%%%%%%%%%%%%%%%%%%%%%%%%%%%%%%%%%%%%%%%%
\linespread{0.9}
\section{Introduction}
\label{sec1}

Robotic systems with embedded artificial intelligence are calling for innovations in design and optimization procedures \cite{khusainov2018automatic}. A multi-disciplinary approach is needed to assimilate models to concurrently optimize the morphology-related Quantities-of-Interest (QoIs) and the decision processes that encapsulate system behavior of robots \cite{danthala2018robotic}. 

% Physical Embodiment of Swarm Systems and Co-design: It is well established that “embodiment" plays a fundamental role in natural intelligence [95], since “morphology facilitates perception, control and communication". For example, the modality and spatio-temporal resolution of the sensory cues available to the agent is strongly embedded in the choice and placement of sensors, and in the locomotion characteristics. Higher resolution sensors often demand greater computing capacities, and greater observability across the swarm demands high-quality communication. Now, choosing powerful computing and radio nodes has space and power consumption implications that require careful consideration. Collective search performance is also directly dependent on the range/speed capabilities embodied by the morphology. Thus, in the context of learning swarm behavior, there are two potentially detrimental consequences of using fixed morphology or ad hoc designs – poor balance in the complexity of sensory, motor and computing systems, resulting in uninteresting collective system dynamics and highly constrained micro-behavior space [95]. There exists work on co-evolving robot morphologies and controllers [96–98], typically for very simple modular organisms with limited functionality [99–104]. In contrast, in this paper, we propose to focus on real-world application capable multi-robot systems comprising small unmanned aerial vehicles or UAVs.

It is well known that ``embodiment" plays a fundamental role in natural intelligence \cite{pfeifer2004embodied}, since \textit{morphology facilitates perception, control and communication}. For example, the modality and spatio-temporal resolution of the sensory cues available to the agent is strongly embedded in the choice and placement of sensors, and in the locomotion characteristics. Higher resolution sensors often demand greater computing capacities, and greater observability across the swarm demands high-quality communication. Now, choosing powerful computing and radio nodes has space and power consumption implications that require careful consideration. %, especially given the small ($\mu$m-cm scale) form factors of emerging swarm robots.
Collective search performance is also directly dependent on the range/speed capabilities embodied by the morphology. %We can conclude that, consideration of morphology encompasses various performance trade-offs and constraints that impact the collective behavior.
Thus, in the context of learning swarm behavior, there are two potentially detrimental consequences of using fixed morphology or ad hoc designs -- poor balance in the complexity of sensory, motor and computing systems, resulting in i) uninteresting collective system dynamics and ii) highly constrained micro-behavior space \cite{pfeifer2004embodied}.

There exists work on co-evolving robot morphologies and controllers \cite{sims1994evolving,lipson2000automatic,lund2003co}, typically for very simple modular organisms with limited functionality \cite{weel2014robotic,khazanov2013exploiting,cheney2013unshackling,komosinski2009evolving,bongard2011morphological,hornby2003generative}. In contrast, in this paper, we propose to focus on multi-robot systems comprising sufficiently complex unmanned aerial vehicles or UAVs that perform real-world operations.

With regards to general frameworks for co-optimizing behavior and morphology of single robots, notable examples include the work by Censi \cite{censi2015mathematical} and Zeng et al \cite{zeng2022an}. The latter proposed an ``artificial life" \cite{bedau2003artificial,belk2020artificial} inspired co-design framework for concurrent design of the system morphology and behavior learning architecture. Behjat et al \cite{behjat2020concurrent} contributed a further advancement of the morphology/learning co-design philosophy by introducing the ``talent metrics" as the parametric space that connects morphology and learning, with the goal to avoid avoiding fully nested co-optimizations. 
% The talent-based co-design framework boasts high optimality (comparable to nested co-design methods) as well as low computing cost (similar to sequential co-design methods). 
This paper builds upon this ``talent metrics" concept by developing a new sequential talent-based co-design framework, and demonstrates its effectiveness with application to co-design UAVs that perform collective signal source location via the algorithm Bayes-Swarm \cite{ghassemi2022penalized}. 

%%%%%%%%%%%%%%%%%%%%%%%%%%%%%%%%%%%%%%%%%%%%%%%%%%%%%%%%%%%%%%%%%%%%%%%%%%%%%%%%%%%%%%%%%%%%%%%%%%%%%%%
% \subsection{Swarm search}

% Robots can be used to navigate many different types of disasters; they can help find a gas-leak in a home \cite{baetz2009mobile}, find stuck climbers after an avalanche \cite{azzollini2021uavbased}, or help victims navigate a nuclear melt down \cite{mohanty2021path}. These are all examples of jobs too dangerous for other humans to attempt without external help or signals \cite{Hamann2018}. 
Usually, robot swarm-search algorithms focus on finding a source signal \cite{Hamann2018} such as a cell phone signal or gas leak, or as large as a nuclear/chemical spill. However, robot swarms face technical challenges such as decentralized communication \cite{wen2018swarm}, independent decision under environmental hazards \cite{shan2020collective}, task allocation between team members \cite{ferrier2012informatics}, and the dynamics of an agent in the context of its environment \cite{zheng2020experimental}. 
% Not to mention the cost of prototyping and simulating the swarm experience \cite{hilder2014pi}. There are algorithms which serve to solve one or more of these issues, but there is no free lunch (NFL). In other words, no one algorithm that can solve all the issues. 
Along with tailoring behavior, it is pivotal to find a set of supportive physical design parameters that enable agents to survive and operate in these generalized environments \cite{afzal2020study}. 

Bayes-Swarm is a recent swarm search algorithm that has been reported to provide superior efficiency compared to other similar algorithms \cite{ghassemi2020extended,ghassemi2022penalized}. Along with providing a decentralized and asynchronous planning process, Bayes-Swarm utilizes both a dynamic active radius around each robot to mitigate conflicts, and a mission-time based heuristic to balance exploration/exploitation of the given environment \cite{rickert2008balancing}. 
% The algorithm uses a batch optimization to figure out the path and Gaussian process to map the environment.  
% It is assumed that all agents are equipped with precise localization  and communication modules. 
% The working of the algorithm is discussed more in Section \ref{Bayesswarm}. 
Parameter tuning plays a major role in the performance of Bayes-Swarm with regards to the above-described capabilities. While hand-crafted adaptive tuning helps, it can not only be tedious to implement but also does not ensure that the tuning process (aka behavior adaptation) is coherent with the physical capabilities of swarm agents (UAVs in our case). %In time restricted missions, the swarm must decide when they'll have enough data from exploring to take an exploitative approach. Though the authors have experimented with these parameters on simulations, there is room for further improvements to make them succeed in more demanding scenarios. 

%%%%%%%%%%%%%%%%%%%%%%%%%%%%%%%%%%%%%%%%%%%%%%%%%%%%%%%%%%%%%%%%%%%%%%%%%%%%%%%%%%%%%%%%%%%%%%%%%%%%%%%
\subsection{Research objectives}

The objective of this paper is to develop a sequential talent-based co-design framework for optimizing the morphology and collective behavior model (CBM) of UAVs performing swarm search based on the asynchronized Bayes-Swarm algorithm. The talent metrics represent operational specifications that are dependent on morphology, and are derived using a set of logical principles. The allowable space of these talent metrics are identified via a multi-objective optimization process, yielding a Pareto front that is used as a constraint in the subsequent process of co-optimizing talent and Bayes-Swarm hyper-parameters. The sequential co-design framework is evaluated over different instances of signal environment. The co-designed performance is then compared with that of a state-of-the-art UAV configuration that uses the baseline Bayes-Swarm algorithm. 

The remaining portion of this paper is organized as follows: Section \ref{sec2} defines the co-design problem and introduces the talent metrics in a generalized form. Section \ref{sec3} elaborates the case study of signal source localization with UAVs using Bayes-Swarm search algorithm. Section \ref{sec4} presents the outcomes of the talent-based co-design case study, and we end with concluding remarks.

%%%%%%%%%%%%%%%%%%%%%%%%%%%%%%%%%%%%%%%%%%%%%%%%%%%%%%%%%%%%%%%%%%%%%%%%%%%%%%%%%%%%%%%%%%%%%%%%%%%%%%% %%%%%%%%%%%%%%%%%%%%%%%%%%%%%%%%%%%%%%%%%%%%%%%%%%%%%%%%%%%%%%%%%%%%%%%%%%%%%%%%%%%%%%%%%%%%%%%%%%%%%%%
\section{Talent-Based Co-design}
\label{sec2}

\begin{figure*}
    \centering
    % \vspace*{5pt}
    \includegraphics[width=0.75\textwidth]{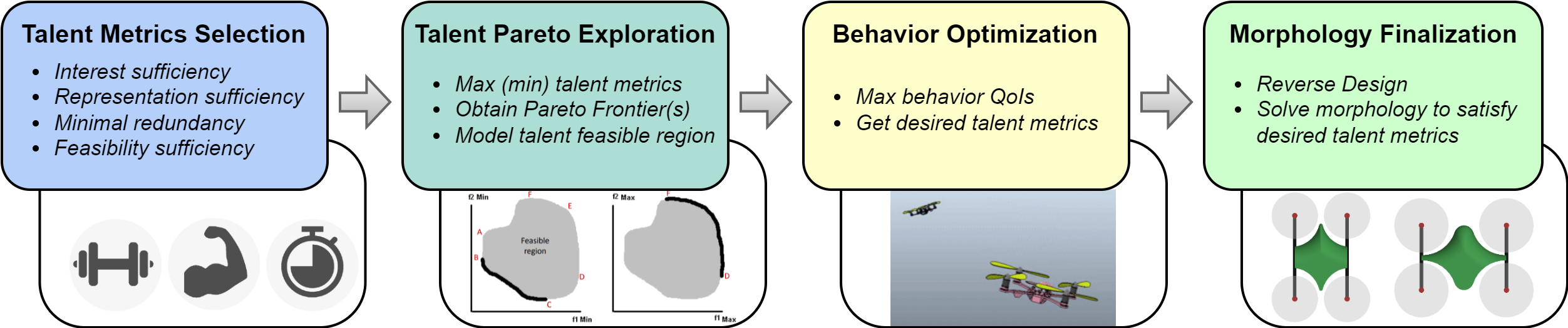}
    \caption{Procedures of the Talent-Based Co-Design Framework}
  \label{fg:flowchart}
\end{figure*}

The non-nested talented-based morphology/CBM co-design framework is presented in this section, along with a generic formulation of typical morphology/learning co-design problems. The proposed framework boasts high optimality (comparable to nested co-design methods) as well as low computing cost (similar to sequential co-design methods). Figure \ref{fg:flowchart} shows the overall procedures of the proposed co-design framework.

%%%%%%%%%%%%%%%%%%%%%%%%%%%%%%%%%%%%%%%%%%%%%%%%%%%%%%%%%%%%%%%%%%%%%%%%%%%%%%
\subsection{Generalized problem formulation}

As proposed by Zeng et al \cite{zeng2022an}, assume a learning problem defined as a state-action pair:
\begin{equation}
\begin{aligned}
    & \mathbf{A}=\mathbf{A}(\mathbf{S},\boldsymbol\Phi,\boldsymbol\Theta) \\
    & \mathbf{S}'=\mathbf{S}'(\mathbf{A},\mathbf{X}_M)
\end{aligned}
\label{EQ:learning_generic}
\end{equation}
where $\mathbf{A}$ represents the set of actions, $\mathbf{S}$ or $\mathbf{S}'$ represent the set of states/scenarios, $\boldsymbol\Phi$ represents the hyperparameters of the learning structure (example: type and size of a neural network), $\boldsymbol\Theta$ represents the learnt parameters (example: weights and biases of a neural network), and $\mathbf{X}_M$ represents the morphology of the physical plant.

Then, the objective of learning is defined as the following:
\begin{equation}
    f_L=f_L(\mathbf{X}_M,\boldsymbol\Phi)=\sum_s^\mathbf{S}R(S,A)
    \label{EQ:objfun_learn}
\end{equation}
where $R$ represents the episodic reward as the result of action $A$ upon state $S$ ($S \in \mathbf{S}$, $A \in \mathbf{A}$). 

Meanwhile, the objective of morphology-dependent Quantities-of-Interest (QoI) is defined as the following:
\begin{equation}
    f_M=f_M(\mathbf{X}_M)
    \label{EQ:objfun_morph}
\end{equation}

The morphology $\mathbf{X}_M$ being the mutual contributor to both $f_L$ and $f_M$ creates couplings in the design space and constraints. On the one hand, morphology parameters favored by learning are misaligned with those favored by performance QoIs. On the other hand, generic morphological constraints that hinder the learnt behavior are almost always present. 

%%%%%%%%%%%%%%%%%%%%%%%%%%%%%%%%%%%%%%%%%%%%%%%%%%%%%%%%%%%%%%%%%%%%%%%%%%%%%%%%%%%%%%%%%%%%%%%%%%%%%%%

\subsection{The talent metrics}

Observed by Behjat et al \cite{behjat2020concurrent}, the morphology($\mathbf{X}_M)$)-to-collective-behavior($f_L$) mapping is indirect and can be highly coupled, hence the introduction of the \textbf{Talent Metrics} (denoted as $\mathbf{Y}_{\texttt{TL}}=\mathbf{Y}_{\texttt{TL}}(\mathbf{X}_M)$). The talent metrics are defined as: morphology-dependent Quantities-of-Interest that (potentially) contribute to the range of CBM QoIs ($f_L$). 

Accordingly, the left side of Equation \ref{EQ:objfun_learn} can be written as:
\begin{equation}
    f_L=f_L(\mathbf{Y}_{\texttt{TL}},\boldsymbol\Phi)
    \label{EQ:objfun_learn_talent}
\end{equation}
Substituting $\mathbf{X}_M$ with $\mathbf{Y}_{\texttt{TL}}$ brings the following benefits:

\begin{enumerate}
    \item Coincidence between morphology QoIs and talent metrics ($\mathbf{Y}_{\texttt{TL}} \cup f_M$). 
    The morphology QoIs often superset the talent metrics ($\mathbf{Y}_{\texttt{TL}} \in f_M$). If not, obtaining the remaining $\mathbf{Y}_{\texttt{TL}}$ in addition to $f_M$ usually imposes little-to-no additional cost.
    
    \item Nested morphology QoI evaluations avoidable. 
    Once the boundaries and constraints of $\mathbf{Y}_{\texttt{TL}}$ are established as a prior, evaluating the CBM QoIs ($f_L$) on longer needs evaluations of $\mathbf{Y}_{\texttt{TL}}$.
    
    \item Likely dimension reduction through CBM QoI evaluations. 
    Typically, the dimension of $\mathbf{X}_M$ is considerably larger than that of $\mathbf{Y}_{\texttt{TL}}$. Dimension reduction helps decreasing computing cost on CBM QoI evaluations.
\end{enumerate}

Selection of the talent metrics follow these principles:

\begin{enumerate}
\item \textbf{Interest sufficiency}: The collection of talent metrics sufficiently represents all vital physical properties corresponding to the QoIs and constraints (morphology and collective behavior alike). 

% \item The physical properties represented by the capacity variables should be \textbf{monotonically} correlated to the design objectives and/or constraints. Satisfying this condition allows us to largely ignore reverse design issues during co-optimization.  \ftnote{Not really absolutely necessary, it depends on how we pick the design candidates for the learning stage. }

\item \textbf{Representation sufficiency}: The collection of talent metrics sufficiently represents the contributions from all of the design variables. 

% \item The capacity variable space should not exceed the design variable space. That is, the range and constraints of the capacity variables should be limited so that no infeasible solution is present.

\item \textbf{Minimal redundancy}: Once a set of talent metrics satisfy all the principles mentioned above, additional metrics are redundant and unnecessary.

\item \textbf{Feasibility sufficiency}: The feasible region of the talent metrics set is dictated by the feasible region of the morphology design, e.g. a feasible set of talent metrics must be achievable with a feasible morphology design candidate.
\end{enumerate}

An understanding of the design space is thus useful in selecting the talent metrics. It is preferred that the talent metrics maintain positive correlations to the CBM QoIs.

%%%%%%%%%%%%%%%%%%%%%%%%%%%%%%%%%%%%%%%%%%%%%%%%%%%%%%%%%%%%%%%%%%%%%%%%%%%%%%%%%%%%%%%%%%%%%%%%%%%%%%%

\subsection{Talent-based co-design framework}

As shown in Figure \ref{fg:flowchart}, the talent-based co-design framework comprises of 4 procedures: 1) talent metrics selection, 2) talent exploration (morphology optimization), 3) optimization of the CBM, and 4) morphology finalization. Talent metrics selection is introduced in the previous subsection.

The talent exploration is done through optimizations of the morphology designs ($\mathbf{X}_M$) to explore the feasible region of the talent metrics ($\mathbf{Y}_{\texttt{TL}}$), defined as the following optimization problem:
\begin{equation} 
\begin{aligned}
&\textrm{Max:} & & f_M=f_M(\mathbf{X}_M) \\
&\textrm{Subject to:} & &\mathbf{X}_M \in \mathbf{R} \ \text{(real numbers)} \\
& & &\mathbf{X}_{\texttt{min}} \leq \mathbf{X}_M \leq \mathbf{X}_{\texttt{max}} \\
& & &g( \mathbf{X}_M ), \ h( \mathbf{X}_M ) \ \text{(if present)}\\
% & & & = 0 \ \text{(if present)}\\
\end{aligned}
\label{eq:objfun_morph_gen}
\end{equation}
Note that $\mathbf{Y}_{\texttt{TL}} \in f_M$. The above optimization exposes the upper limits of the talent metrics. If a negative talent-to-collective-behavior correlation is likely, the minimization optimization of Equation \ref{eq:objfun_morph_gen} is also necessary to expose the lower limits of the talent metrics.

A model representing the limitations of the talent metrics (Pareto Frontiers) is created:
\begin{equation}
    \hat{Y}_{\texttt{TL},1} = f_{\text{SM}} \left(\hat{Y}_{\texttt{TL},2},\ldots,\hat{Y}_{\texttt{TL},m} \right)
    \label{eq:talent_sm}
\end{equation}
where $f_{\text{SM}}$ refers to the multivariate surrogate model with the inputs of $\left(\hat{Y}_{\texttt{TL},2},\ldots,\hat{Y}_{\texttt{TL},m} \right)$ and the output of $\hat{Y}_{\texttt{TL},1}$. $\hat{Y}_{\texttt{TL},i}$ represents the $i$th dimension of a non-dominated (e.g. on Pareto Frontier) point of talent metrics, $m$ is the dimension of $\hat{Y}_{\texttt{TL}}$. The talent metrics surrogate model maps the upper limits of the talents.

The collective behavior model optimization performs a standalone optimization of maximizing the collective behavior QoIs, defined as the following:
\begin{equation}
\begin{aligned}
&\textrm{Max:} & & f_{L}=f_{L}(\mathbf{Y}_{\texttt{TL}},\boldsymbol\Phi) \\
&\textrm{Subject to:} & & \mathbf{Y}_{\texttt{min}} \leq \mathbf{Y}_{\texttt{TL}} \leq \mathbf{Y}_{\texttt{min}} \\
& & &g_1( \mathbf{Y}_{\texttt{TL}} ): \ Y_{\texttt{TL},1}-f_{\text{SM}} \left(Y_{\texttt{TL},2},\ldots,Y_{\texttt{TL},m} \right)\leq 0 \\
& & &g( \boldsymbol\Phi ) , h( \boldsymbol\Phi ) \ \text{(if present)}\\
\end{aligned}
\label{eq:objfun_learn_gen}
\end{equation}
The constraint $g_1$ verifies the feasibility of proposed talent metrics by checking whether the proposed $\mathbf{Y}_{\texttt{TL},1}$ is on or behind the modeled Pareto Frontier $f_{\text{SM}} \left(Y_{\texttt{TL},2},\ldots,Y_{\texttt{TL},m} \right)$ (assume $\mathbf{Y}_{\texttt{TL}}$ is positively correlated to $f_{L}$). Constraint $g_1$ ensures the feasibility of optimized talent metrics, aka the existence of a satisfying morphology design. 

The morphology finalization is achieved through another optimization with the morphology design:
\begin{equation}
\begin{aligned}
&\textrm{Min:} & & f_f=||\mathbf{Y}_{\texttt{TL}}(\mathbf{X}_M)-\mathbf{Y}_{\texttt{TL}}^*|| \\
&\textrm{Subject to:} & &\mathbf{X}_M \in \mathbf{R} \\
& & &\mathbf{X}_{\texttt{min}} \leq \mathbf{X}_M \leq \mathbf{X}_{\texttt{min}} \\
& & &g( \mathbf{X}_M ), \ h( \mathbf{X}_M ) \ \text{(if present)}\\
% & & & = 0 \ \text{(if present)}\\
\end{aligned}
\label{eq:objfun_final_gen}
\end{equation}
where $f_f$ refers to the objective function value, and $\mathbf{Y}_{\texttt{TL}}^*$ represents the optimal talent metrics obtained in procedure 3) the CBM optimization ($\mathbf{Y}_{\texttt{TL}}^* = \textrm{argmax} \ f_{L}(\mathbf{Y}_{\texttt{TL}},\boldsymbol\Phi)$). Equation \ref{eq:objfun_final_gen} searches for morphology design candidates satisfying the desired talent metrics. Multiple satisfactory design candidates are possible.

%%%%%%%%%%%%%%%%%%%%%%%%%%%%%%%%%%%%%%%%%%%%%%%%%%%%%%%%%%%%%%%%%%%%%%%%%%%%%%%%%%%%%%%%%%%%%%%%%%%%%%% %%%%%%%%%%%%%%%%%%%%%%%%%%%%%%%%%%%%%%%%%%%%%%%%%%%%%%%%%%%%%%%%%%%%%%%%%%%%%%%%%%%%%%%%%%%%%%%%%%%%%%%
\section{Asynchronized UAV search Swarm}
\label{sec3}

\subsection{The collective behavior model (CBM)} \label{Bayes-Swarm}
%%%%%%%%%%%%%%%%%%%%%%%%%%%%%%%%%%%%%%%%%%%%%%%%%
The Bayes-Swarm algorithm utilized in this paper is adopted from Ghassemi et al. \cite{ghassemi2020extended,ghassemi2022penalized}, with its original implementation available as a GitHub repository \cite{PayamghaBayesSwarmP}.
%\footnote{\url{https://github.com/adamslab-ub/Bayes-Swarm-P}}.
% The algorithm focuses on finding the source location in complex signal environments using a Gaussian Process. It is a decentralized and asynchronous algorithm, which 
Bayes-Swarm takes advantage of a temporal heuristic to balance explorative and exploitative path planning modes mid-flight. A novel batch-BO acquisition function (Eq. \ref{eq:obj-func}) is formulated to balance exploration and exploitation. 
% The function is expressed in . 
Each robot from the location $x$ takes an action that leads to $x_r$  in a way that maximizes this function. Here the term $\Omega_r$ leads the robot $r$ to the signal source thereby promoting exploitation. The term $\Sigma_r$ minimizes the knowledge uncertainty and pushes the robot $r$ to explore further. The local penalization factor $\Gamma_r$ makes sure the robot $r$ takes a trajectory, or decision horizon which does not conflict with that of their peers. $\beta$ is a scaling parameter that aligns the orders of magnitude between exploration and exploitation. 
% The formulations and detailed explanations can be found in the original paper \cite{ghassemi2020extended}.

% The image \ref{fig:Bayeswarm_acquisition_func} explains the applications of each term.
\begin{equation}\label{eq:obj-func}
    \begin{aligned}
  &x_r^{k_r+1} = \underset{x \in X^{K_r}}{arg \max}(\alpha \Omega_r + (1 - \alpha)\beta \Sigma_r)\Gamma_r \\
     &\text{s.t.\hspace{10pt}}
0 \leq l_s^{k_r} = ||x-x_r^{k_r}|| \leq VT
    \end{aligned}
\end{equation}
where, 
\begin{equation} \label{eq:alpha_new}
    \begin{aligned}
        \alpha = 1/(1+\exp(-a(\frac{t}{T_{max}}- b)))
    \end{aligned}
\end{equation}

\begin{figure}
     \centering
    %  \vspace*{7pt}
% 
        \includegraphics[width=0.75\linewidth]{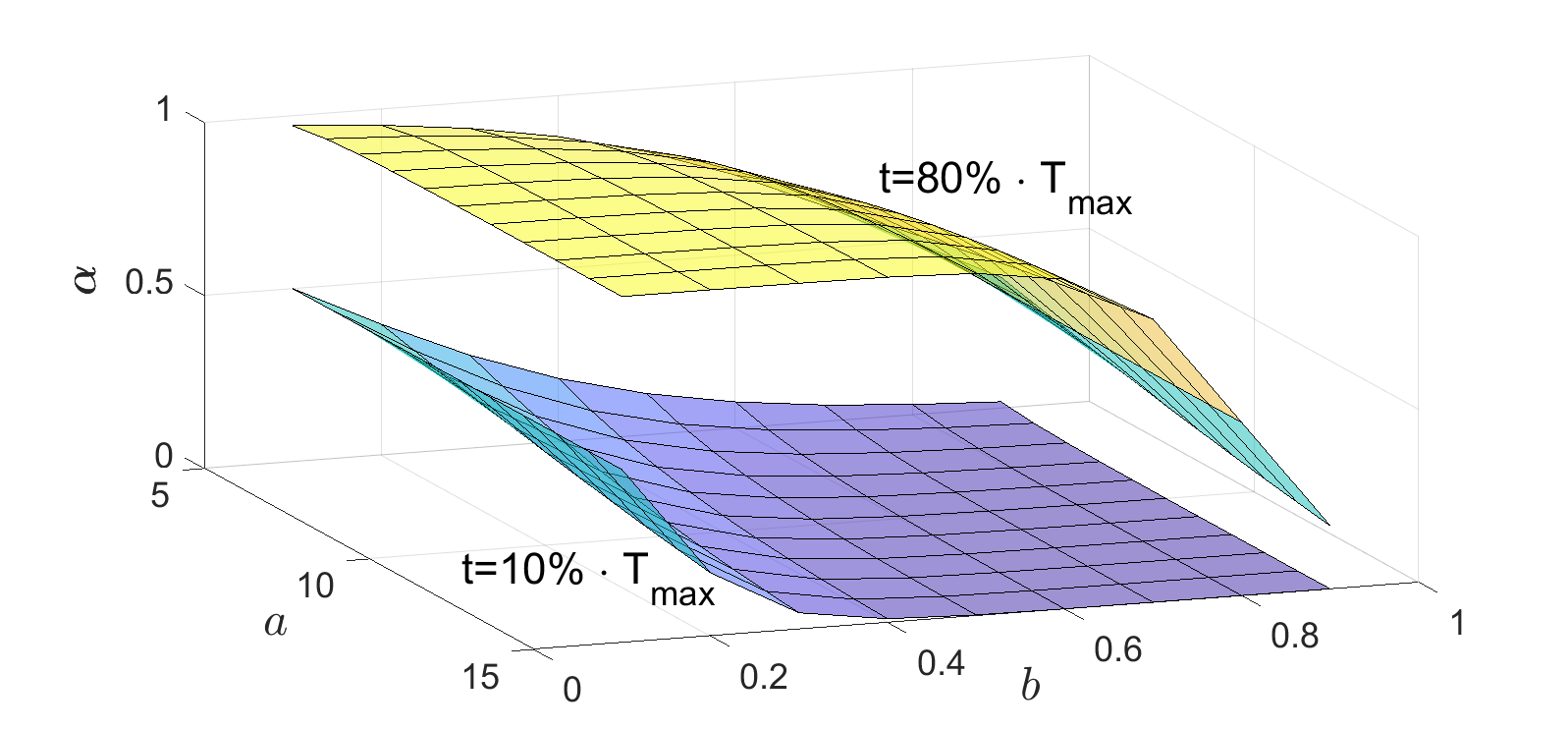}
        \caption{$\alpha$ Response Surfaces}
        \label{fig:alpha_curves}
\end{figure}

\begin{figure}
     \centering
        \includegraphics[width=0.6\linewidth]{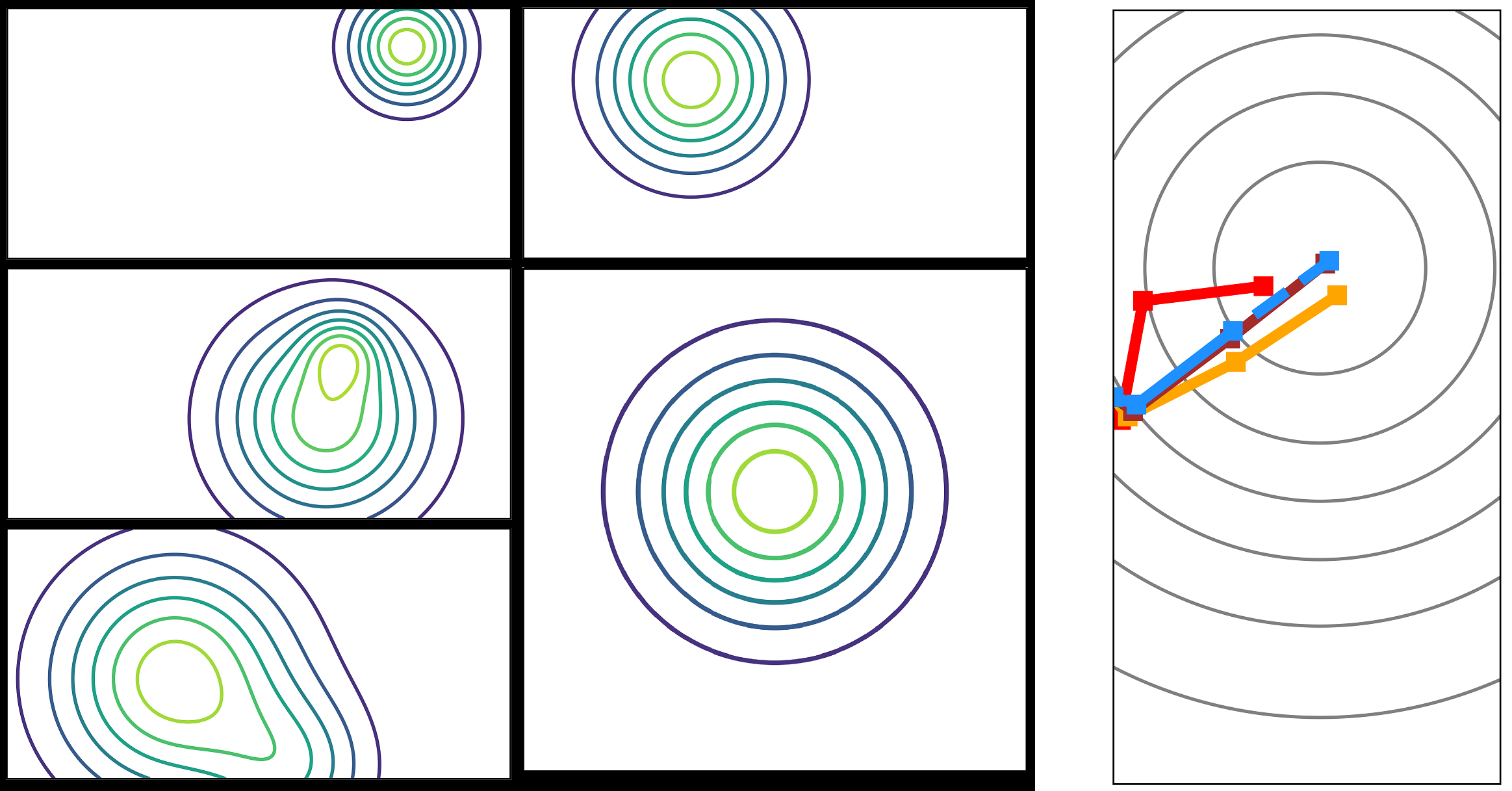}
        \caption{Some variations of test environments (left), example of search swarm trajectories (right).}
        \label{fig:different_sources}
\end{figure}

The heuristic $\alpha \in [0,1]$ determines the exploitation weight and is determined by Equation \ref{eq:alpha_new}, where $t$ is the current mission time and $T_{max}$ is the maximum allotted time for the mission. Due to this, $\alpha$ is adaptive and modelled in such a way that the robots are purely explorative in the beginning and gradually changes to exploitative as the mission time approaches $T_{max}$. Figure \ref{fig:alpha_curves} shows the response surfaces of $\alpha$ at different times.
% The flowchart of the Bayes-Swarm algorithm can be found in Figure \ref{fig:Bayeswarm_flowchart}. 
It is important to note that the robot swarm is asynchronous, which means each agent can take information and act independently if necessary, without waiting for their peers to relay information. Some of the environments with different signal types are shown in Figure \ref{fig:different_sources}. The environments have  two different dimensions, (30 * 15) km and  (30 * 30) km. These environments have random source targets and signal strength. 

% \begin{figure}
%      \centering
% %      \begin{subfigure}[t]{0.3\linewidth}
% %          \centering
% %          \includegraphics[width=\textwidth]{source11.png}
% % 		\caption*{(a)}
% %      \end{subfigure}
% %      \hfill
%      \begin{subfigure}[t]{0.3\linewidth}
%          \centering
%          \includegraphics[width=\textwidth]{source12.png}
% % 		\caption*{(b)}
%      \end{subfigure}
%      \hfill
%      \begin{subfigure}[t]{0.3\linewidth}
%          \centering
%          \includegraphics[width=\textwidth]{source13.png}
% % 		\caption*{(c)}
%      \end{subfigure}
%      \hfill
%      \begin{subfigure}[b]{0.3\linewidth}
%          \centering
% 		\includegraphics[width=\textwidth]{source14.png}
% % 		\caption*{(d)}
%      \end{subfigure}
%      \hfill
%      \begin{subfigure}[b]{0.3\linewidth}
%          \centering
% 		\includegraphics[width=\textwidth]{source15.png}
% % 		\caption*{(e)}
%      \end{subfigure}
%         \caption{Examples of Variations in Test Environments}
%         \label{fig:different_sources}
% \end{figure}

The CBM objective is to optimize the hyper-parameters in the acquisition function for maximum success rate and low search time.  Here we optimize the  weight function ($\alpha$) and consider 2 hyper-parameters $a$ and $b$, which plays the role of balancing exploration and exploitation ratio. $a$ and $b$ form the CBM design space ($\boldsymbol\Phi$).
% Ideally, $a$ and $b$ shall vary based on the environment. In this paper, $a$ and $b$ are to be optimized for a variety of scenarios.

\begin{figure}[h]
% \vspace{1mm}
\centering
\includegraphics[width=0.5\linewidth]{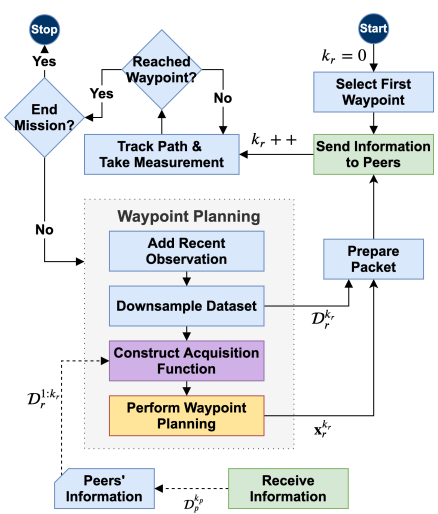}
\caption{Flowchart Diagram of Bayes-Swarm} 
\label{fig:Bayeswarm_flowchart}

\end{figure}

\begin{table}
\begin{center}
% \vspace{1.8mm}
\caption{Stochastic Test Environment Settings}
\label{tab:sample_set}
\scriptsize
    \begin{tabular}{ll}
    \toprule
    Parameter & Value \\
    \midrule
    Sample Set Size & 100 \\
    Sampling Method & Latin Hypercube Sampling \\
    Swarm Size & 6 to 15 \\
    Source Types & 5 \\
    \multirow{2}{*}{Source Location} & -14 to +14 km (E-W) \\
    & 0 to +14 km (N-S) \\
    Signal Strength & 15 to 100 \\
    \bottomrule
    \end{tabular}
\end{center}
% \vspace{-3mm}
\end{table}

%%%%%%%%%%%%%%%%%%%%%%%%%%%%%%%%%%%%%%%%%%%%%%%%%%%%%%%%%%%%%%%%%%%%%%%%%%%%%%%%%%%%%%%%%%%%%%%%%%%%%%%

\subsection{Morphology design and selection of talents}

Our case study features a Blended-Wing-Body (BWB) integrated quadcopter (BIQU) introduced in a previous research \cite{zeng2019artificial}. This quadcopter utilizes an H-shaped frame with an aerodynamic Blended-Wing-Body that helps reducing drags and creating some lift during cruising flight states. 
% The proposed quadcopter design provides a practical lift during cruising flight without creating a significant drag, thus contributes to better endurance. 
% Figure \ref{FG:cadmodel} illustrates the configuration of the aircraft.

Analytical and numerical models are created to estimate the mass, aerodynamics, power-to-thrust correlations, typical cruising speeds, flight endurance, as well as the sensor performance. 

% \begin{figure}
%     \centering
%     \includegraphics[width=0.75\linewidth]{topview.png}
%     \caption{The BWB-integrated quadcopter}
%     \label{FG:cadmodel}
% \end{figure}

Following the principles introduced previously, we choose the 1) typical cruising speed, 2) flight range, and 3) sensor detection distance as the talent metrics. The aircraft flight range can be shorter than the width of the search area. The sensor detection distance dictates how close a search UAV needs to approach the target until a confirmed found.

%%%%%%%%%%%%%%%%%%%%%%%%%%%%%%%%%%%%%%%%%%%%%%%%%%%%%%%%%%%%%%%%%%%%%%%%%%%%%%%%%%%%%%%%%%%%%%%%%%%%%%%

\subsection{The optimization problems}

Adopting the generalized problem formulation presented in the previous section, our case study comprises of 3 optimization problems from 2 design spaces in serial. 

The first optimization problem is for talent exploration:
\begin{equation}
\begin{aligned}
&\textrm{Max:} & & \mathbf{Y}_{\texttt{TL}}=f_M(\mathbf{X}_M) \\
&\textrm{S. t.:} & &\mathbf{X}_M \in \mathbf{R} \\
& & &\mathbf{X}_{\texttt{min}} \leq \mathbf{X}_M \leq \mathbf{X}_{\texttt{max}} \\
\end{aligned}
\label{eq:objfun_morph_1}
\end{equation}
where $\mathbf{Y}_{\texttt{TL}}$ includes the flight range, the typical cruising speed, and the sensor detection distance; $f_M$ represents the outcomes of various models for the BIQU UAV. The talent metrics happen to be identical as the morphology QoIs in this case study.

A surrogate model needs to be trained to capture the Pareto Frontier of $\mathbf{Y}_{\texttt{TL}}$ (Equation \ref{eq:talent_sm}). Considering the captured Pareto Frontier is approximated from a heuristic (NSGA-II results), we choose to train a \textbf{Gaussian process} model to adapt for the imprecise samples. It is better to set less coupled or sensitive variables as inputs and put the opposite as outputs (although not necessary for the complexity for our case study).Therefore, the inputs of the model are the cruising speed and sensor detection distance, the output is the flight range.

The second optimization problem is for the optimization of the CBM:
\begin{equation}
\begin{aligned}
&\textrm{Min:} & & f_{L}=\sum_{i=1}^{100}t_{i}(\mathbf{Y}_{\texttt{TL}}, \boldsymbol\Phi) \\
&\textrm{S. t.:} & & \mathbf{Y}_{\texttt{min}} \leq \mathbf{Y}_{\texttt{TL}} \leq \mathbf{Y}_{\texttt{max}} \\
& & & \boldsymbol\Phi_{\texttt{min}} \leq \boldsymbol\Phi \leq \boldsymbol\Phi_{\texttt{max}} \\
& & & g_1( \mathbf{Y}_{\texttt{TL}} ): \ Y_{\texttt{TL},1}-f_{\text{SM}}  \left(Y_{\texttt{TL},2},Y_{\texttt{TL},3} \right)\leq 0 \\
&\textrm{while:} & & t_{i} = 5 \times 10^4 \ \textrm{if search $i$ fails.} \\ 
\end{aligned}
\label{eq:objfun_learn_1}
\end{equation}
where $t_{i}$ represents the search time cost in case $i$ (100 cases in total, see Table \ref{tab:sample_set}), $\boldsymbol\Phi$ refers to $a$ and $b$ from Equation \ref{eq:alpha_new}, $Y_{\texttt{TL},1}$, $Y_{\texttt{TL},2}$, and $Y_{\texttt{TL},3}$ correspond to the flight range, cruising speed, and the sensor detection distance respectively. 
% The constraint function $C( \mathbf{Y}_{\texttt{TL}} )$ measures the distance between the chosen point of $\mathbf{Y}_{\texttt{TL}}$ and the Pareto Frontier of $\mathbf{Y}_{\texttt{TL}}$ (dominated $\mathbf{Y}_{\texttt{TL}}$ points are feasible nevertheless). 

A significant penalty is applied to $t_{i}$ if the search fails to converge, emphasizing high success rate over short search time. Possible reasons of failure include time cost exceeding limits and stagnated waypoint updating algorithm. Due to the optimizer limitations (MATLAB \texttt{bayesopt}), the talent feasibility constraint $g_1( \mathbf{Y}_{\texttt{TL}} )$ is applied through an external penalty.

The third optimization problem solves the optimized morphology design according to the optimal talent metrics:
\begin{equation}
\begin{aligned}
&\textrm{Min:} & & f_f=||\mathbf{Y}_{\texttt{TL}}(\mathbf{X}_M)-\mathbf{Y}_{\texttt{TL}}^*|| \\
&\textrm{S. t.:} & &\mathbf{X}_M \in \mathbf{R} \\
& & &\mathbf{X}_{\texttt{min}} \leq \mathbf{X}_M \leq \mathbf{X}_{\texttt{min}} \\
\end{aligned}
\label{eq:objfun_final_1}
\end{equation}

Table \ref{tb:DVs} lists the morphology variables, the talent metrics, and their respective boundaries.
\begin{table}
% \vspace{7pt}
	\begin{center}
		\caption{Design Variables and Talent Metrics}
		\label{tb:DVs}
		\scriptsize
		\begin{tabular}{c|ccc}
			\toprule
			Type & Variable & LB & UB \\
			\midrule
			\multirow{6}{*}{$\mathbf{X}_M$}  & Length & 0.2m & 0.5m \\
			& Width & 0.2 m & 0.5 m \\
			& Motor Size & 100 W & 400 W \\
			& Battery Size & 13.9 W$\cdot$h & 55.6 W$\cdot$h \\
			& Propeller Size & 7 inch & 12 inch \\
			& Sensor Weight & 0.1 kg & 0.7 kg \\
			\hdashline
            \multirow{3}{*}{$\mathbf{Y}_{\texttt{TL}}$}  & Flight Range & 8.9 km & 32.6 km \\
			& Cruising Speed & 4.5 m/s & 9.5 m/s \\
			& Sensor Detection Dist. & 100 m & 1000 m \\
			\hdashline
			\multirow{2}{*}{$\Phi$}  & $a$ & 5 & 15 \\
			& $b$ & 0.1 & 0.9 \\
			\bottomrule
		\end{tabular}
	\end{center}
\end{table}

%%%%%%%%%%%%%%%%%%%%%%%%%%%%%%%%%%%%%%%%%%%%%%%%%%%%%%%%%%%%%%%%%%%%%%%%%%%%%%%%%%%%%%%%%%%%%%%%%%%%%%%

\subsection{Configurations and preferences}

The morphology optimizations (talent exploration and morphology finalization) adapt an NSGA-II \cite{NSGA2} solver, while the CBM optimization utilizes a Bayesian Optimization \cite{pelikan1999boa} solver (Matlab \texttt{bayesopt}). The settings are adjusted in the two Case Studies for balancing convergence quality and computing load. Table \ref{tb:config1} summarizes the settings of the respective optimization solvers.

The numerical case study is completed on a Windows desktop computer in MATLAB 2021b. Parallel computing with 10 workers is enabled for all of the optimization trials. 
% Table \ref{tb:config3} summarizes the settings and performance of the optimization trials.

\begin{table}
% \vspace{6pt}
%     \renewcommand\arraystretch{1.0}
% 	\setlength{\abovecaptionskip}{0.cm}
% 	\setlength{\belowcaptionskip}{-0.cm}
	\begin{center}
		\caption{Optimizer Settings}
		\label{tb:config1}
		\scriptsize
		\begin{tabular}{cccccc}
			\toprule
			\multicolumn{6}{c}{NSGA-II} \\
% 			\hdashline
			Repeat & Pop. & Max & X-over & Mutation & Mutation \\
			Runs &  Size  & Iter. & Rate & Rate & Strength \\
			\hdashline
			5 & 200 & 75 & 0.9 & 0.4 & 0.075 \\
			\midrule
			\multicolumn{6}{c}{Bayesian Optimization} \\
% 			\hdashline
			Kernel & Acq. & Seed & Max & GP Active \\
			Func.  & Func. & Points & Iter. & Set & \\
			\hdashline
			Matern    & EI+  & 25 & 150  & 50 & \\
			\bottomrule
		\end{tabular}
	\end{center}
\end{table}

% \begin{table}
% %     \renewcommand\arraystretch{1.0}
% % 	\scriptsize
% % 	\setlength{\abovecaptionskip}{0.cm}
% % 	\setlength{\belowcaptionskip}{-0.cm}
% 	\begin{center}
% 		\caption{Bayesian Optimization Settings}
% 		\label{tb:config2}
% 		\begin{tabular}{ccccc}
% 			\toprule
			 
% 			\bottomrule
% 		\end{tabular}
% 	\end{center}
% \end{table}

% \begin{table}
% %     \renewcommand\arraystretch{1.0}
% % 	\scriptsize
% % 	\setlength{\abovecaptionskip}{0.cm}
% % 	\setlength{\belowcaptionskip}{-0.cm}
% 	\begin{center}
% 		\caption{Workstation Configurations}
% 		\label{tb:config3}
% 		\begin{tabular}{ll}
%          \toprule
%         %  Category & Parameter & Value \\
%         %  \midrule
%           CPU & AMD Ryzen 3700X \\
%           Frequency & 4.10 Ghz \\
%           No. of Cores & 8C16T \\
%           System RAM & 32 GB \\
%           Software & MATLAB R2021b (Windows 10) \\
% 		\bottomrule
% 		\end{tabular}
% 	\end{center}
% \end{table}

%%%%%%%%%%%%%%%%%%%%%%%%%%%%%%%%%%%%%%%%%%%%%%%%%%%%%%%%%%%%%%%%%%%%%%%%%%%%%%%%%%%%%%%%%%%%%%%%%%%%%%% %%%%%%%%%%%%%%%%%%%%%%%%%%%%%%%%%%%%%%%%%%%%%%%%%%%%%%%%%%%%%%%%%%%%%%%%%%%%%%%%%%%%%%%%%%%%%%%%%%%%%%%

\section{Result Analysis}
\label{sec4}

%%%%%%%%%%%%%%%%%%%%%%%%%%%%%%%%%%%%%%%%%%%%%%%%%%%%%%%%%%%%%%%%%%%%%%%%%%%%%%%%%%%%%%%%%%%%%%%%%%%%%%%

\subsection{A baseline design}

To comprehend the quality of the optimized design, a baseline design is proposed based on a popular F-450 modular quadcopter made by DJI. Thanks to the BWB fuselage, the baseline design could achieve a longer flight range than the F-450 quadcopter, despite mostly identical geometries and choices of components. Table \ref{tb:results} lists the morphology, the talent metrics, and performances in swarm searches of the baseline design.

\subsection{The optimized UAV and swarm hyperparameters}

\begin{figure}
    \centering
    \includegraphics[width=0.9\linewidth]{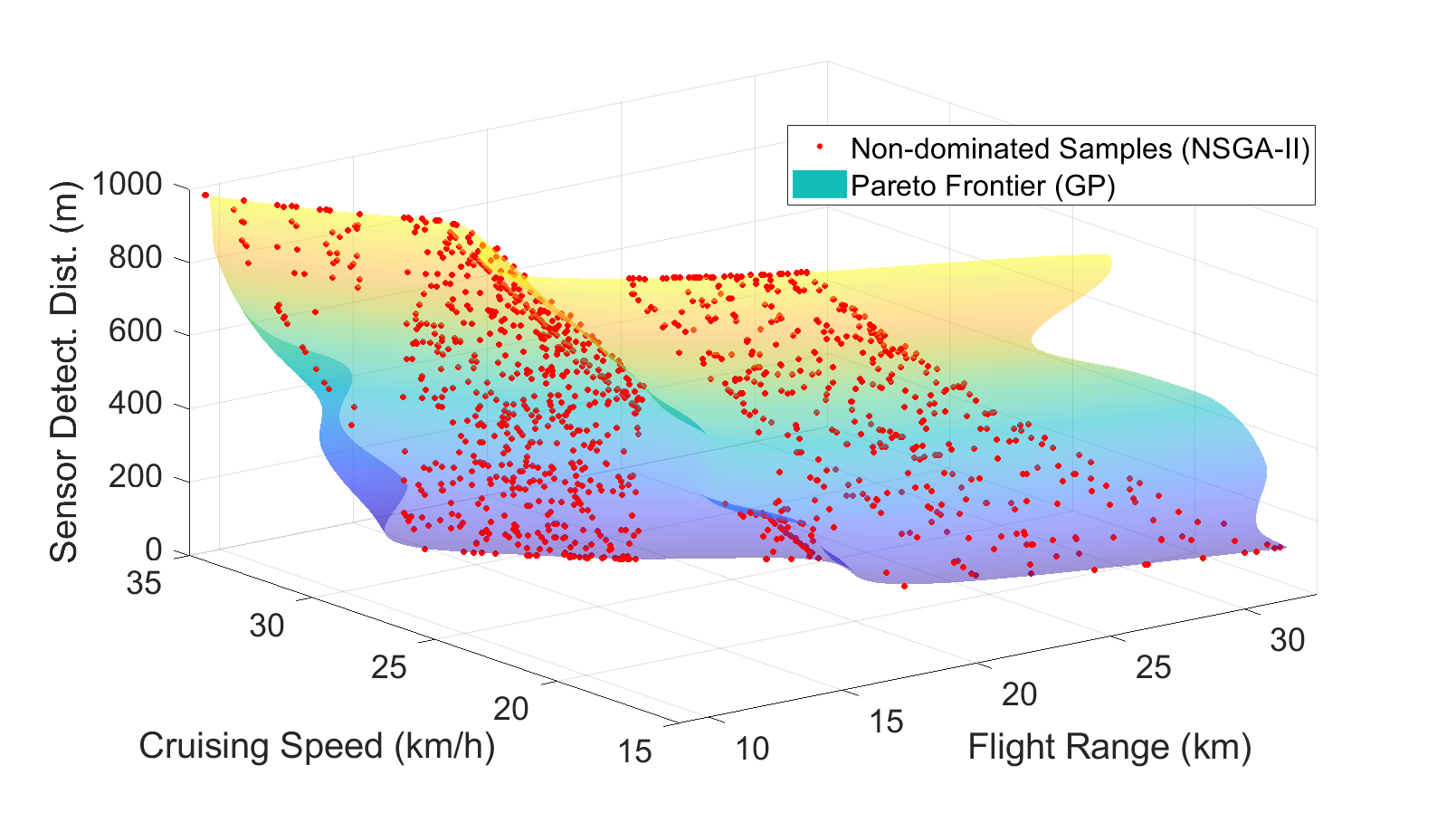}
    \caption{Pareto Frontier of the Talent Metrics (with GP Model)}
    \label{FG:talent_pareto}
\end{figure}

The final populations from the 5 optimization trials for talent exploration are gathered to sort out the selection of non-dominated design candidates. A total of 1313 non-dominated candidates are sorted out to train a Gaussian Process surrogate model representing the Pareto Frontier of the talent metrics (shown in Figure \ref{FG:talent_pareto}).

\begin{figure}
% \vspace{5pt}
    \centering
    \includegraphics[width=1.0\linewidth]{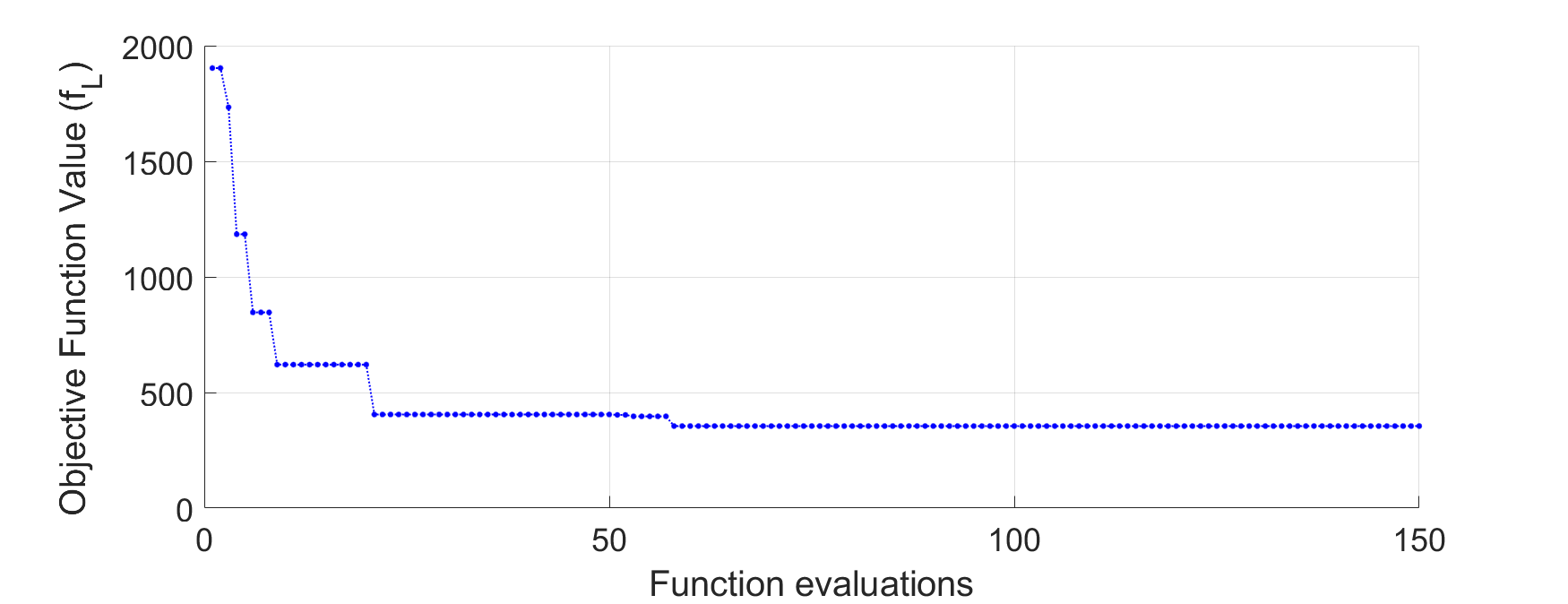}
    \caption{Collective Behavior Model Optimization Convergence History}
    \label{fg:convhist}
\end{figure}

The convergence history of the CBM optimization is shown in Figure \ref{fg:convhist}. The quick reduction in $f_L$ indicates the iterations are sufficient.

From the optimal talent metrics obtained through the CBM optimization, the optimized morphology design is solved. The morphology parameters, Bayes-Swarm hyperparameters, talent metrics, and the average search time of the optimized design are listed in Table \ref{tb:results}. 

\begin{table}
	\begin{center}
		\caption{Parameters: Final Design V.S. Baseline}
		\label{tb:results}
		\scriptsize
		\begin{tabular}{c|c|c|c}
			\toprule
			Type & Variable & Final Design & Baseline \\
			\midrule
			\multirow{6}{*}{$\mathbf{X}_M$}  & Length & 0.315 m & 0.325 m \\
			& Width & 0.315 m & 0.325 m \\
			& Motor Size & 100 W & 175 W \\
			& Battery Size & 55.6 W$\cdot$h & 55.6 W$\cdot$h \\
			& Propeller Size & 12 inch & 9.7 inch \\
			& Sensor Weight & 150 g & 240 g \\
			\hdashline
            \multirow{3}{*}{$\mathbf{Y}_{\texttt{TL}}$} & Flight Range & 24.0 km & 16.7 km \\
			 & Cruising Speed & 4.7 m/s & 5.1 m/s \\
			& Sensor Detection Dist. & 100 m & 400 m \\
			\hdashline
			\multirow{2}{*}{$\Phi$}  & $a$ & 5.66 & 10.0 \\
			& $b$ & 0.788 & 0.50 \\
			\hdashline
			\multirow{2}{*}{$f_L$} & Avg. Search Time & 2764 s & 2197 s \\
			 & Success Rate & 92.6\% & 75.8\% \\
			\bottomrule
		\end{tabular}
	\end{center}
\end{table}

% \begin{table}
% 	\begin{center}
% 		\caption{Performance Improvements Over Optimization}
% 		\label{tb:results}
% 		\small
% 		\begin{tabular}{c|c|c|c}
% 			\toprule
% 			UAV Design & CBM & Avg. Search Time & Success Rate \\
% 			\midrule
% 			Baseline & Random Search & 12,838 s & 78\% \\
% 			Baseline & Bayes-Swarm & 2242 s & 78\% \\
% 			Optimal & Random Search & 2493 s & 74\% \\
% 			Optimal & Bayes-Swarm & 2764 s & 92.6\% \\
% 			\bottomrule
% 		\end{tabular}
% 	\end{center}
% \end{table}

\begin{figure}
    \centering
    \includegraphics[width=0.6\linewidth]{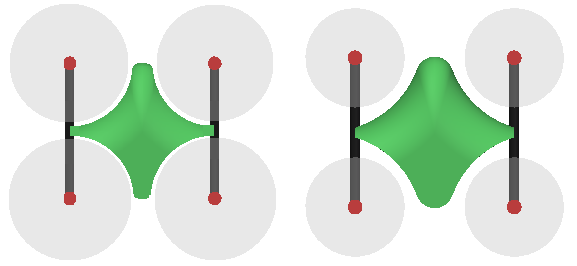}
    \caption{Optimized UAV (left) and the Baseline UAV (right)}
    \label{fg:optmshape}
\end{figure}

Compared to the baseline design, our optimized design increase the success rate from 76\% down to 92\%, while the average search time of 2197 seconds is longer than the 2764 seconds from the baseline. 
Figure \ref{fg:optmshape} illustrates subtle differences in UAV shape between the optimized design and the baseline. Coincidentally, our final design is pretty similar to the baseline in terms of size and cruising speed, while the major differences are on the flight range, detection distance, and CBM hyperparameters. 

\begin{figure}
% \vspace{5pt}
    \centering
    \includegraphics[width=0.85\linewidth]{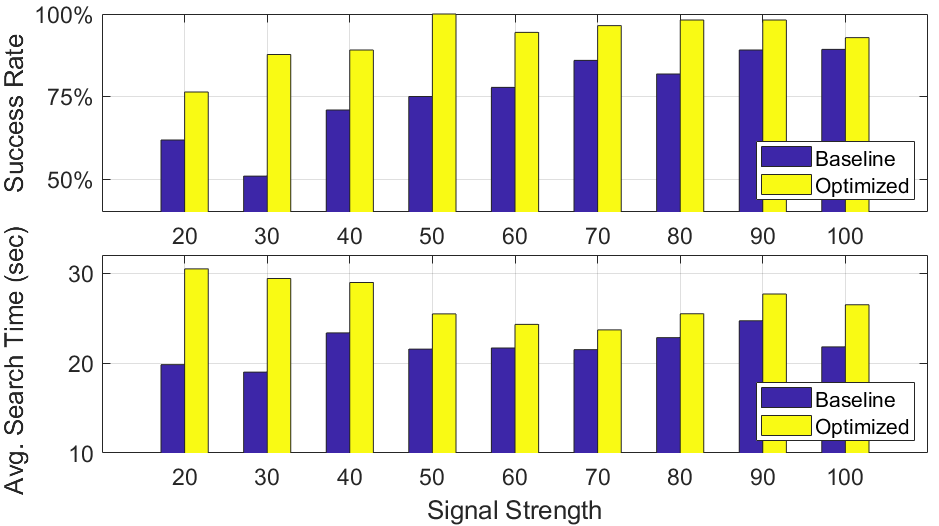}
    \caption{Swarm Performance w.r.t. Signal Strength}
    \label{fg:barplot_versus}
\end{figure}

Observing the optimized talent metrics and swarm hyperparameters, we can conclude the long flight range is more desirable than fast flight speed and long sensor range when it comes to maximizing the success rate. Figure \ref{fg:barplot_versus} indicates the optimized design maintains high success rate even if the source signal is weak (less than 50).
% The optimized final morphology/collective behavior combination shows significant advantage in mission robustness and efficiency.

%%%%%%%%%%%%%%%%%%%%%%%%%%%%%%%%%%%%%%%%%%%%%%%%%%%%%%%%%%%%%%%%%%%%%%%%%%%%%%%%%%%%%%%%%%%%%%%%%%%%%%%

\subsection{Computing costs analysis}

The 5 optimization trials for talent exploration took 32 minutes 6 seconds (1926 seconds), with a self time (cost of the algorithm) of 13 minutes 13 seconds; the Bayesian optimization trial for the CBM optimization took 3 hours 29 minutes 36 seconds (12,576 seconds), with a self time of 26 seconds; the optimization trial for morphology design costs 6 minutes 52 seconds; modeling the feasible region of the talent metrics costs less than 30 seconds to compute. The overall computing time of the entire case study is just over 4 hours (247 minutes).

The average computing cost of morphology evaluation is negligible (around 0.12 second), the average cost of evaluating the CBM objective is around 83.0 seconds (with parallel computing). Based on the computing costs observed from the trials, the hypothetical computing cost of running a fully nested co-design optimization with the NSGA-II solver with identical settings is estimated as 14.6 days (350 hours). In comparison, our talent-based co-design optimization is extremely frugal on computing load.
    
%%%%%%%%%%%%%%%%%%%%%%%%%%%%%%%%%%%%%%%%%%%%%%%%%%%%%%%%%%%%%%%%%%%%%%%%%%%%%%%%%%%%%%%%%%%%%%%%%%%%%%% %%%%%%%%%%%%%%%%%%%%%%%%%%%%%%%%%%%%%%%%%%%%%%%%%%%%%%%%%%%%%%%%%%%%%%%%%%%%%%%%%%%%%%%%%%%%%%%%%%%%%%%

\section{Concluding Remarks}
\label{sec5}
This paper developed a co-design approach for maximizing the performance of teams of UAVs performing signal source search over a variety of signal environments and team sizes. This approach decomposes the concurrent morphology/behavior optimization process into a sequence of three sub-optimizations, which respectively \textbf{1)} obtain the best trade-offs among morphologically-regulated capabilities aka ``talent metrics", \textbf{2)} use them to constrain the talent variables during the process of optimizing the hyper-parameters of the search behavior model, and \textbf{3)} use the best trade-off or Pareto front and the behavior optimization outcome to further optimize and finalize the morphology of the UAVs. To implement this co-design framework, we use the swarm search algorithm called Bayes-Swarm. Three talent metrics, namely cruising speed, flight range and sensor detection range, are selected following distinct principles related to morphology-behavior relationships. The co-optimization process finished in $\sim$4 hours, resulting in roughly (estimated) two orders of magnitude reduction of computing time compared to a naive nested morphology-behavior optimization process. While the optimized UAV design resulting from our co-design framework takes slightly longer search completion times compared to a baseline design (over successful scenarios), the success rates of the former are much higher -- this is expected as the penalized objective function driving the behavior optimization process is designed to strongly favor search success over search time. 

Confined by computing power, our case study only utilized the maximization of the talent metrics, which could possibly lead to falsely feasible designs. Additional numerical experiments are planned in the immediate future to investigate the necessity and benefits of capturing the full range of the talent metrics. The test environments of the numerical experiments are kept deliberately simple, i.e., they do not include real-world complexities such as obstacles and more meaningful sensor features, in order to highlight the fundamental algorithmic contributions (of co-design) in this paper and keep the overall simulation costs manageable. Such assumptions can be alleviated in the future, in order to transition these capabilities to field experiments. %Lastly, the current implementation searches for fixed optimum values of the swarm (collective behavior) algorithm hyper-parameters, thereby leaving the opportunity to explore mechanisms that can optimally adapt the behavior model to team size and apriori known features of the operational environment.  %but the Bayes-Swarm search can be adapted to reflex real-world obstacles by modifying the environment map. We are also working to implement widely-adapted benchmark case studies (like openAI environments) and more challenging applications (UAV learning to fly from scratch, for example) to widen the use cases of the talent-based co-design framework. 

% \section*{Acknowledgement}
% This work was supported by the National Science Foundation (NSF) award CMMI 2048020. Any opinions, findings, conclusions, or recommendations expressed in this paper are those of the authors and do not necessarily reflect the views of the NSF.
 
\bibliographystyle{IEEEtran}
\bibliography{IEEEabrv,IEEEexample,bibliography}

\end{document}